# Image Classification and Optimized Image Reproduction

[1]Jaswinder Singh Dilawari , [2]Dr. Ravinder Khanna

[1] Ph.D Research Scholar, Pacific University, Udaipur, Rajasthan, INDIA
dilawari.jaswinder@gmail.com

[2] Principal, Sachdeva Engineering College for Girls, Mohali, Punjab, INDIA
ravikh_2006@yahoo.com

**Abstract**

By taking into account the properties and limitations of the human visual system, images can be more efficiently compressed, colors more accurately reproduced, prints better rendered. To show all these advantages in this paper new adapted color charts have been created based on technical and visual image category analysis. A number of tests have been carried out using extreme images with their key information strictly in dark and light areas. It was shown that the image categorization using the adapted color charts improves the analysis of relevant image information with regard to both the image gradation and the detail reproduction. The images with key information in hi-key areas were also test printed using the adapted color charts.

**Key Words**: *ICC-profiles,RGB,Gray Scales,Color Test Chart*

## 1. Introduction

ICC-profiles are being used more and more frequently to predict the rendering of colors and thereby ensure a high quality. An ICC-profile is a data file describing the color characteristics of an imaging device (Sharma, 2004). The primary purpose or use of this file is to maintain color consistency in images viewed, displayed or printed on various devices (Wallner, 2000). By using a common format (ICC, International Color Consortium) for characterization of color units, it is easier to determine the color gamut of a device and thereby optimize a print-out. A device is characterized by printing and measuring target values in a color chart. There are a large number of different color charts on the market, all of which are assumed to be valid for all types of images, no matter whether the relevant image information is located in high-key areas, low-key areas or mid-tones.

The result is that too few color tones containing key image information can be analyzed. In the work described in Paper I, new adapted color charts were created based on technical and visual image category analysis. A number of tests have been carried out using extreme images with their key information strictly in the dark or light areas. The results show that the image categorization using the adapted color charts improves the analysis of relevant image information with regard to both image gradation and detail reproduction. The new adapted color charts preserve details in the low-key areas, and give a more distinct image with a better fidelity to the original image. Evaluations have been made using a test panel and the pair-comparison method.

## 2. OBJECTIVE

The purpose of this work was to study if the categorization of images can improve the quality of color reproduction by adapting standard color charts.

## 3. METHOD

Test charts commonly used for output characterization were studied to evaluate how the tone steps are distributed for output characterization, and a new set of color values was used to create an image-adapted test chart, different from the gamma and gradation values normally used. These category-adapted test charts were printed under controlled conditions. Spectral measurements were made on the new test charts, and new output profiles were calculated and applied in the RGB-to-CMYK conversion for the specific image category aimed for. A validation print was made with the new separation values applied to the specific image category aimed for. The results were evaluated by the subjective pair-comparison method, where 50 people with a graphic arts background judged the result. An objective evaluation was made by instrument measurements of lightness values.

3.1 Background to the creation of an image category border
- Classification using L*-values

In order to establish more distinct borders between the different image categories, tests were carried out with from each category were selected together with Björn Olsson (who introduced Swedish definitions for different image categories), Figure 1.

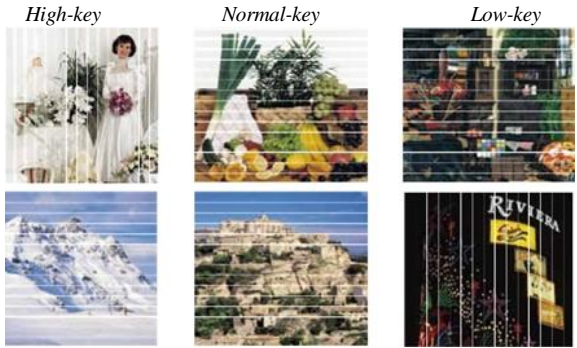

Fig 1. Examples of a high-key, a normal-key and a low-key image. (Royalty free images from Stockpix - the lower row)

The distribution of the L*-value in the three types of images indicates the borders that mayexist between these images, Figure 2.

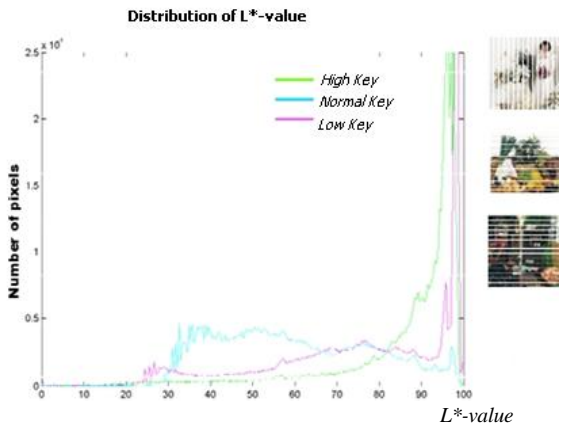

Fig 2. Distribution of L*-value for high-key, normal-key and low-key images. The peak for high L-values in the normal-key image is caused by the white background. The graph was created in Matlab (Enoksson, 2001).

The images were processed in Adobe Photoshop, where the color information was discarded in order to analyze only the L*-values. The images as well as their histograms were studied and analyzed using the Matlab-software, Figure 3.

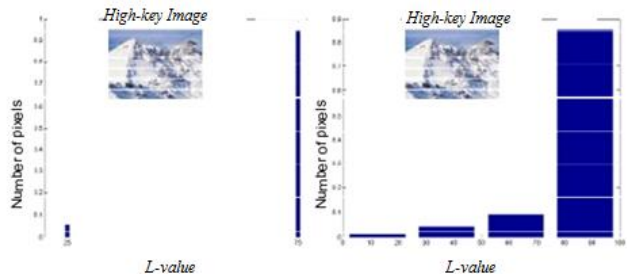

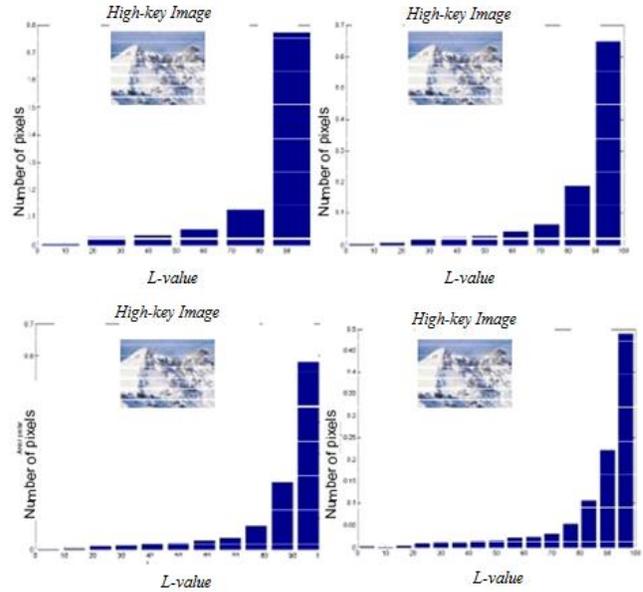

Fig.3. The steps made in the Matlab analysis. Number of pixels in different steps of L* scale in high-key image. The steps made it possible to find the borders between the images.

The borders of the L*-values for the different image types were used to create an image-adapted color chart.

Each image category was printed in an offset press, Heidelberg Speedmaster 74, on a coated (130g/m$^2$) and an uncoated (130g/m$^2$) paper. The prints were processed and separated using Adobe Photoshop, where the color information in the images was compressed against adjacent colors, turning them into an IT.8-target (24x18 patches) for easier measurement, Figure 4.

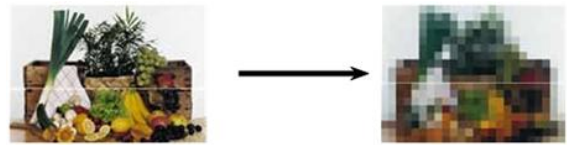

Fig 4. Adapting of the image (normal-key) to an IT.8 arget, 24x18 patches.

The patches of the images were measured using a spectrophotometer and the CIE Lab L*-value was computed. The measured values were used to compare the three image categories the L*-values in the original data and the L*-values on coated and uncoated papers, Figure 5. The Figure clearly shows that the scatter of the L*-values in the images was compressed by the different paper grades. On an uncoated paper, the dark areas are clearly lighter, which means that there is a poorer detail rendering in the printing.

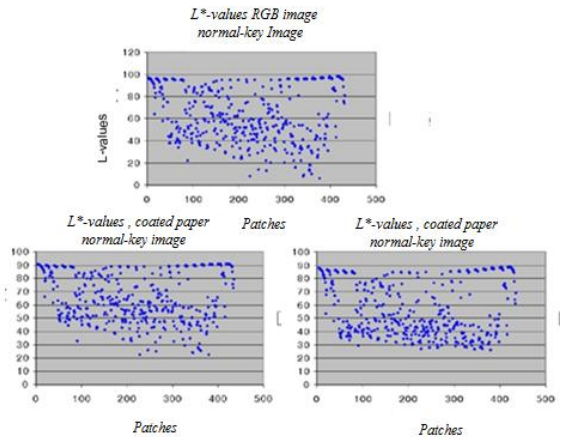

Fig 5. Distribution of L*-values for a normal-key image and for prints on coated and uncoated paper.

## 4. RESULTS

### 4.1 Image classification by using L-values

The studies of the three image categories (high-key, low-key, normal-key) revealed that the borders in the L*-scale for high-key images were 100-60, for normal-key images 60-40, and for low-key images 40-0, Figure 6.

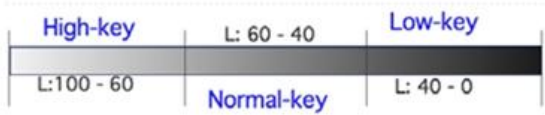

Fig 6. The borders in the L*-scale (Lightness) for high-key, normal-key and low-key images.

### 4.2. The image adapting of the test chart

Are there other solutions which will make it easier to give priority to interesting areas than is usually done by tonal compression or optimal separation (GCR - Gray component Replacement, UCR - Under Color Removal)? The beginning and the end of the production chain both offer an adaptation to the production and method of giving priority to certain image categories and areas of an image.

Each part of the graphics industry has high demands on color reproduction and the demands of the print buyers and end users for quality are steadily increasing. Col-
or communication between scanners, computers and output devices has improved,thanks to ICC-profiles. There are several companies developing software for profiles
on the market. Each of these products is designed to help the user achieve improved color fidelity, each one looks and works differently and may produce different results (Adams II, 2000). Each software has its own color test chart. Test charts differ from each other in the number of color patches, the values of the patches and the color distribution.

The test charts have one thing in common - they are intended to work for any kind of images with no focus on any particular image category.

The hypothesis in this Paper I has been that it is possible to adapt the test chart to the image category and thus give priority to sections of the tonal range. Tests of this hypothesis have revealed that there are two ways to adapt the test chart:

a) to create a new adapted test chart
b) to adapt the standard test chart

a) Creation of a new adapted test chart

The borders suggested in Figure 6 were used to create a new image-adapted test chart, as shown in Figure 7. For some patches, the Neugebauer equations have been used.

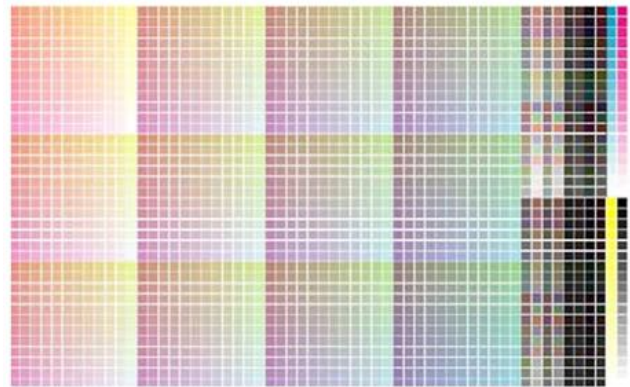

Fig 7. An example of the new image-adapted test chart.

The construction of the test charts available on the market was studied and the values of these test charts were measured and compared (Paper I). The new test charts were created based on the suggested borders between image tone values, Figure 32. The distribution of the values generates a slope which can be compared to a gamma curve for the different image types (Paper I).

b) Adapting of the standard test chart

Another way of adapting the test charts is to adapt the standard test chart. The same
knowledge about the gradation from the previous study was used. The the standard test chart 6.02 was adapted in software AdobePhotoshop (Paper I).

### 4.3 Adapting the IT8 test chart for printing

A new printing was carried out (Heidelberg Speedmaster 74) using these test charts and subjective and objective evaluation of the prints were carried out:

- the subjective evaluation used 50 people from the graphic industry and from the Graphic Institute. A paired comparison (Bristow, Johansson, 1983) was made of the prints. People involved in this evaluation preferred the prints which were based on separations with the adapted test charts, Figure 8.

- for the objective evaluation, gray scales were created in Adobe Photoshop and separated with the same profiles as for the images. The gray scale which was based on a separation with the adapted test chart showed more detail in the dark tones.Subjective evaluation: low-key images

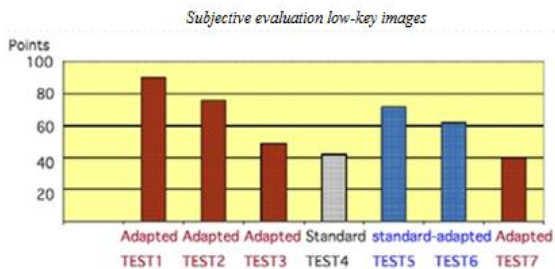

Fig 8. The result of the subjective evaluation. The y-axis shows the points for the test charts

The low-key prints which were based on a separation with the image-adapted test chart showed more detail in the dark areas, as can be seen in Figure 9.

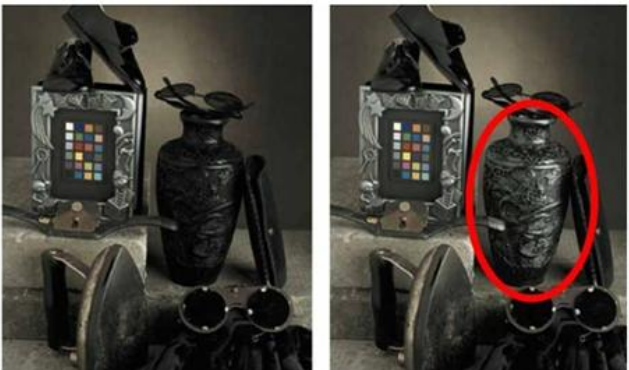

Fig 9. A comparison between low-key images after printing. The image to the left was separated with the standard test chart and the image to the right with the image-adapted test chart.

## CONCLUSION

The results suggest that an adjustment only to low-key images is sufficient, as even normal-key prints then show a better fidelity to the original image. High-key images show no difference between the different IT.8 test charts, Figure 39. Classifying images is a difficult task, as it is the customer who should ultimately decide which areas of an image are most important. The general reasoning that high-key images have the greatest concentration of information in the bright areas, normal-key images in the middle-tone areas, and low-key images in the dark areas of the tone scale is quite reasonable. An "exact" mathematical definition can be produced, but it loses its value directly for the graphics industry (because of the different needs of the users) as it does not help in the actual image processing. An analysis of the pixel numbers in an image in the L*-scale generates suggested borders which can be applied in further studies. These borders make it possible to adapt the IT.8 test chart for printing with improved results. IT.8 test charts for scanners also permit a certain adaptation for production or for specific colors.

The ideas and methods concerning the adaptation of IT.8 test charts is the subject of a application, where a Swedish patent has already been granted. (Enoksson, 2004). No similar work was found by the patent office.

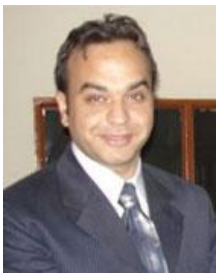

**Jaswinder Singh Dilawari** is a Ph.D Research Scholar, Pacific University, Udaipur, Rajasthan, INDIA and is working as an Associate Professor , Geeta Engineering College, Panipat, Haryana ,India .He has teaching experience of 12 years .His area of interest includes Computer Graphics, Computer Architecture ,Software Engineering ,Fuzzy Logic  and Artificial Intelligence .He is life member of  Indian Society for Technical Education (ISTE)

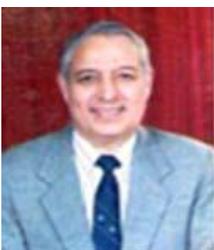

**Dr.Ravinder Khanna** Graduated in Electrical Engineering from Indian Institute of Technology(IIT) Dehli in 1970 and Completed hisMasters and Ph.D degree in Eletronicsand Communications Engineering fromthe same Institute in 1981 and 1990respectively. He worked as an ElectronicsEngineer in Indian Defense Forces for 24Years where he wasinvolved in teaching,research and project magement of someof the high tech weapon systems. Since 1996 he has full timeSwitched to academics. he has worked in many premiere technicalinstitute in india and abroad. Currently he is the Principal ofSachdeva Engineering College for Girls, Mohali, Punjab (India).He is active in the general area of Computer Networks, ImageProcessing and Natural Language Processing